\documentclass[10pt,twocolumn,letterpaper]{article}

\usepackage{cvpr}              

\usepackage{graphicx}
\usepackage{amsmath}
\usepackage{amssymb}
\usepackage{booktabs}
\usepackage{xcolor}         
\usepackage[accsupp]{axessibility}

\usepackage[pagebackref,breaklinks,colorlinks]{hyperref}

\usepackage[capitalize]{cleveref}
\crefname{section}{Sec.}{Secs.}
\Crefname{section}{Section}{Sections}
\Crefname{table}{Table}{Tables}
\crefname{table}{Tab.}{Tabs.}

\usepackage{amsmath}
\DeclareMathOperator{\relu}{ReLU}

\DeclareMathOperator{\gelu}{GELU}
\DeclareMathOperator{\swish}{Swish}
\DeclareMathOperator{\mish}{Mish}
\DeclareMathOperator{\acts}{Act_{S}}
\DeclareMathOperator{\erf}{erf}
\DeclareMathOperator{\sigmoid}{sigmoid}
\DeclareMathOperator{\softplus}{softplus}

\usepackage{listings}
\definecolor{codegreen}{rgb}{0,0.6,0}
\definecolor{codegray}{rgb}{0.5,0.5,0.5}
\definecolor{codepurple}{rgb}{0.58,0,0.82}
\definecolor{codebackcolour}{rgb}{0.95,0.95,0.92} 

\lstdefinestyle{mystyle}{
	backgroundcolor=\color{codebackcolour},   
	commentstyle=\color{codegreen},
	keywordstyle=\color{magenta},
	numberstyle=\tiny\color{codegray},
	stringstyle=\color{codepurple},
	basicstyle=\ttfamily\footnotesize,
	breakatwhitespace=false,         
	breaklines=true,                 
	captionpos=b,                    
	keepspaces=true,                 
	numbers=left,                    
	numbersep=5pt,                  
	showspaces=false,                
	showstringspaces=false,
	showtabs=false,                  
	tabsize=2
}

\crefformat{footnote}{#2\footnotemark[#1]#3}

\def\cvprPaperID{97} 
\def\confName{CVPR}
\def\confYear{2022}

\begin{document}

\title{PEA: Improving the Performance of ReLU Networks for Free by Using Progressive Ensemble Activations}

\author{\'Akos Utasi\\
Continental AI Development Center, Budapest\\
{\tt\small akos.utasi@continental-corporation.com}
}
\maketitle

\begin{abstract}
In recent years novel activation functions have been proposed to improve the performance of neural networks, and they show superior performance compared to the ReLU counterpart. However, there are environments, where the availability of complex activations is limited, and usually only the ReLU is supported. In this paper we propose methods that can be used to improve the performance of ReLU networks by using these efficient novel activations during model training. More specifically, we propose ensemble activations that are composed of the ReLU and one of these novel activations. Furthermore, the coefficients of the ensemble are neither fixed nor learned, but are progressively updated during the training process in a way that by the end of the training only the ReLU activations remain active in the network and the other activations can be removed. This means that in inference time the network contains ReLU activations only. We perform extensive evaluations on the ImageNet classification task using various compact network architectures and various novel activation functions. Results show 0.2--0.8\% top-1 accuracy gain, which confirms the applicability of the proposed methods. Furthermore, we demonstrate the proposed methods on semantic segmentation and we boost the performance of a compact segmentation network by 0.34\% mIOU on the Cityscapes dataset.
\end{abstract}

\section{Introduction}
\label{sec:intro}

Currently, the most widely used activation function in neural networks is the Rectified Linear Unit (ReLU) \cite{Nair2010}. However, in the past couple of years both hand-crafted (\eg Gaussian Error Linear Unit - GELU \cite{Hendrycks2016}, or Mish \cite{Misra2020}), and search-based (Swish \cite{Ramachandran2018}) smooth, non-monotonic, continuous activation functions have been proposed, which outperform ReLU networks. In general, these activations provide a better information propagation in the network, and they solve some weaknesses of ReLU, for example the problem of dying ReLUs. Currently, networks using novel activations can be found at the top of the leaderboards of popular benchmarks such as ImageNet \cite{Deng2009} classification, or MS-COCO \cite{Lin2014} object detection. Although these activations can be used to improve the performance, in some environments they have no hardware support, which limits their applicability. For example in many embedded systems (\eg in autonomous driving systems) the hardware supports the ReLU (or optionally some ReLU-variant, \eg ReLU6) activation only, and hence using these novel activations in such systems is cumbersome or, in the worst case, not feasible.

In this paper, we propose training methods to improve the performance of ReLU networks. We use state-of-the-art smooth, non-monotonic, continuous activation functions (referred as SOTA activation hereinafter) in the beginning of the training process, and as the training progresses the network is adapted to the ReLU activation. In the proposed methods an ensemble is created from these two activations. Moreover, the activation coefficients of the ensemble are progressively updated during training in a way that the coefficient of the SOTA activation is decayed towards zero. We name our training method as Progressive Ensemble Activations or PEA. At the end of the PEA training process we obtain a network containing ReLU activations only, and the network can be deployed in environments, where only the ReLU activation is supported. We propose two ensemble variants: i) a weighted ensemble model linearly combining the two activations, and ii) a stochastic variant, where the activation function is randomly selected. Our experiments show that the proposed methods improve the accuracy of the resulting ReLU network. Note that the ensemble is used during training only, but the network used during inference is a ReLU network. This means that from the network's point of view the improvement is \emph{for free}, as the network architecture during inference is unchanged. In this work, our main focus is on improving compact, low-latency networks, that can be used in real-time vision applications \eg in mobile or embedded systems.

We extensively evaluate the proposed methods on the ImageNet classification task in \cref{sec:exp_class}, using compact ReLU networks and various recent SOTA activations. According to our experiments, the proposed PEA training process results in ReLU networks with improved top-1 accuracy, and this improvement can go as high as 0.82\%. Although our main focus is on compact networks, we also applied PEA for training a large ResNet-50 network, and improved its top-1 accuracy by 0.23\%.
Next, we demonstrate the proposed methods on semantic segmentation task in \cref{sec:exp_semseg}, and we improved the performance of a compact neural network by 0.34\% mIOU on the Cityscapes dataset.

Our contributions are summarized as follows:
\begin{enumerate}
  \item We propose the PEA training method to improve the performance of ReLU networks. PEA uses ensemble activations during training and updates the ensemble coefficients progressively, while in inference time the ensemble activation is replaced by a ReLU.
  \item We propose two ensemble variants: i) in the weighted ensemble the activations are linearly combined, and ii) in the stochastic ensemble the activations are randomly selected.
  \item We demonstrate the capabilities of the proposed method on ImageNet classification. In our experiments we mainly focus on compact low-latency networks (AlexNet, MobileNet, ERFNet, ResNet-18). In addition, we also apply the method on a larger ResNet-50 network. The proposed methods improved the top-1 accuracy of these networks by 0.2--0.8\%.
  \item We demonstrate the proposed method on semantic segmentation, and we improve the performance of the compact ERFNet network by 0.34\% mIOU on the Cityscapes dataset.
\end{enumerate} 


\section{Related Works}

In our proposed methods we use SOTA activation functions during the network training, hence we start with a brief introduction to some of the recent activations, especially those that showed great success on popular vision benchmarks. Then we discuss techniques that can be used for improving the accuracy, and more specifically we focus on techniques, that do not modify the inference time network architecture. For simplicity, we refer to these techniques as \emph{for-free} methods, which in general improve the accuracy with modifications in the training process, but the final model structure used during inference is left intact.

\subsection{Activation Functions}\label{sec:activations}

Early neural networks used $\sigmoid$ or $\tanh$ as activation, but they were ineffective in deep neural networks. The Rectified Linear Unit (ReLU) \cite{Nair2010} was the first successful activation in deep neural networks \cite{Glorot2011}, and is still the most widely used activation in various networks. Moreover, the ReLU is supported by a large range of hardware, including GPUs, TPUs, mobile or embedded AI accelerators. However, it has some well-known weaknesses, such as dying ReLUs \cite{Lu2020}, which can be mitigated by using careful weight initialization \cite{He2015}, regularization during the training (\eg Dropout \cite{Srivastava2014}), or normalization techniques (\eg batch normalization \cite{Ioffe2015}) in the network architecture. 

To overcome these weaknesses, some other works replace the ReLU with more efficient activation functions, such as Leaky-ReLU \cite{Maas2013}, Exponential Linear Unit (ELU) \cite{Clevert2016}, or Scaled Exponential Linear Unit (SELU) \cite{Klambauer2017}. In recent state-of-the-art neural networks the ReLU is usually replaced by smooth, non-monotonic, continuous activations, which retain the advantageous properties of ReLU, \eg they are bounded below resulting in sparsity, and unbounded above resulting in unsaturated output. On the other hand, they result in a smooth output landscape and hence in a better gradient flow. Moreover, they also reduce the dying neurons phenomenon, and unlike the ReLU they are continuously differentiable. One of the earliest such activation is the Gaussian Error Linear Unit (GELU) \cite{Hendrycks2016}, defined as $\gelu\left(x\right)=x\cdot \frac{1}{2}\left[1+\erf\left(x/\sqrt{2}\right)\right]$. With the recent introduction of Transformers for vision tasks \eg \cite{Dosovitskiy2021,Liu2021} the GELU activation has become very popular. The Swish activation was proposed in \cite{Ramachandran2018}, and unlike previous hand-crafted activations, it was discovered by architecture search. Swish is defined as $\swish\left(x\right)=x\cdot\sigmoid\left(\beta x\right)$, where $\beta$ is either a constant or a trainable parameter, and if $\beta=1$, it is equivalent to the Sigmoid-weighted Linear Unit (SiLU) \cite{Elfwing2018}. In their extensive experiments they found that Swish consistently outperforms ReLU in most cases. Finally, the Mish activation, defined as $\mish\left(x\right)=x\cdot \tanh\left(\softplus\left(x\right)\right)$, was proposed in \cite{Misra2020}. Its design was inspired by the self-gating mechanism of Swish, which means the input is multiplied with the non-linear function of the input. In their evaluations they found that Mish outperforms other activations, including GELU and Swish. Note that these activations are relatively complex, but efficient implementations are available for high-end GPUs, \eg Swish is supported in cuDNN 8.2.

The superior performance of these novel activation functions is also apparent in popular vision benchmarks outperforming previous ReLU networks. For example, at the time of writing the manuscript the transformer-based ViT-G/14 network was leading the ImageNet benchmark with 90.94\% top-1 accuracy, and it uses GELU activations. The best convolutional neural networks include EfficientNet-L2 using Swish (top-1=90.2\%), and NFNet-F4+ using GELU (top-1=89.2\%). The top of the MS-COCO object detection \cite{Lin2014} leaderboard was populated by Transformer based networks, which use the GELU activation. Such works include \cite{Zhang2022, Liu2022, Yuan2021} where the backbone is a Swin Transformer \cite{Liu2021} variant, and their average precision (AP) is in the range of $62.4-63.3$. The best performing convolutional detector ($AP=57.3$) \cite{Ghiasi2021} is a Cascade Mask-RCNN, based on the EfficientNet-B7 architecture, which uses the Swish activation. Nevertheless, the applicability of neural networks using these activations is limited in resource-limited systems, such as embedded AI accelerators, where only the ReLU activation is supported by the hardware.

\subsection{For-Free Methods}

As discussed above in \cref{sec:activations} the accuracy of neural networks can be improved by using one of the novel activation functions proposed recently.
However, the improvements of this approach are \emph{not for-free}, because i) these activations come with an increased inference time caused by their complex formulation, and ii) in the worst case they have lack of support in several environments, such as embedded AI accelerators.

On the other hand, there are existing approaches that can be used to improve the performance of neural networks \emph{for-free}. Such widely used approaches for convolutional neural networks include different random image augmentations (pixel manipulation, or geometric transformation) \eg RandAugment \cite{Cubuk2020}, or regularization techniques such as weight decay \cite{Krizhevsky2012}, or Dropout \cite{Srivastava2014}. Some other works improve the performance by using a better optimization method to train the network, and i) replace the vanilla SGD with advanced algorithms such as the adaptive Adam \cite{Kingma2015} or RAdam \cite{Liu2020}, or the Lookahead mechanism proposed in \cite{Zhang2019}, or ii) use hand-crafted learning rate schedule, such as piece-wise constants, exponential/polynomial/cosine decay, or using restart techniques \cite{Loshchilov2017}.
The Stochastic Depth method \cite{Huang2016} is specially designed for improving deep ResNet networks, and during training it randomly disables the residual branch of a block, bypassing it through the identity transform in the skip connection branch. The Stochastic Weight Averaging (SWA) method \cite{Podoprikhin2018} creates and ensemble in the weight space, and combines the weights of the network at different stages of the training into a single model, which is obtained by maintaining the running average of these weights. In our experiments we also use some of these well-known techniques, in particular we use simple data augmentations, dropout, weight decay, different learning rate schedules, and adaptive optimizers. 

Recently, the ExpandNets method \cite{Guo2020} was proposed to improve the performance of networks, and it applies linear over-parameterization during training. Each layer is expanded into a succession of multiple layers without using any non-linearity between them. This results in an increased number of network parameters during training, but due to the lack of non-linear transformations they can be contracted back into the original form after the training is finished.
Although the method does not affect the inference time, the additional matrix operations in the expansion results in significantly increased training time and memory footprint, \eg the authors reported a 2 to 5 times slower training of compact networks using an expansion rate of 4, which can be a limitation for training larger networks.

Another approach for improving compact networks is knowledge distillation (KD) where a trained teacher (typically a big network) provides knowledge either i) as the supervisory signal \cite{Hinton2015}, or ii) as some regularization of intermediate layers of the compact student network \cite{Romero2015,Komodakis2017}. However, recent works in KD \cite{Cho2019, Mirzadeh2020, Shen2020, Shen2021} found that the final accuracy of the student is largely affected by many different factors of the teacher-student setup, including \eg initialization, label smoothing, weight decay, or teacher quality. Finally, recent advances in self-supervised and semi-supervised learning show promising results, but these methods require a large number of additional unlabeled, but curated training data to improve the performance. For example in \cite{Pham2021} the authors use 300 million unlabeled images to boost the ImageNet top-1 accuracy of their network.


In this paper we propose an alternative \emph{for-free} approach, which is complementary to other techniques discussed above, and its training-time computation overhead is relatively small. To the best of our knowledge there is no prior work that uses SOTA activation functions during training time to improve the performance of ReLU networks. In the proposed method we create ensembles composed of the ReLU and the efficient activation functions discussed in \cref{sec:activations}. During training the ensemble is progressively updated in a way that by the end of the training only the ReLU remains active, and the other activation can be removed. This means that the computation overhead during training is mainly due to the additional point-wise activations, which requires significantly less computation and memory as compared to \eg the additional matrix operations used in ExpandNet. This means that our proposed methods can also be applied to improve larger networks.
 
\section{Proposed Methods}

During training the proposed PEA methods use ensemble activations, which are composed of the ReLU and one of the SOTA activations discussed in \cref{sec:activations}. Let $x\in\mathbb{R}^{n}$ denote a tensor, and let $f\left(x,\alpha\right)$ denote the ensemble activation parameterized by coefficient $\alpha$. We perform model training in three phases, such that by varying the ensemble coefficients $\alpha$ we control the contribution of each activation. In the first part of the training, referred as \emph{initial phase}, only the SOTA activation contributes to the ensemble, and the ReLU is disabled. Next, in a \emph{transition phase} the ensemble is slowly adapted towards the ReLU activation, which means that the contribution of the SOTA activation is decayed, while the contribution of the ReLU is increased. The training process is finalized in the \emph{final phase}, where only the ReLU activation is used to fine-tune the network and the SOTA activations are completely disabled. This means that the final network contains ReLU activations only. In the next subsections we present two different ensemble variants. Note that both methods involve the computation of two activations, but this extra operation is relatively cheap as compared to the other computation-heavy operations like matrix multiplication used in convolutions or in fully-connected layers.

\subsection{Weighted Ensemble}\label{sec:lin_ens}

\begin{figure}
	\centering
	\includegraphics[width=0.9\linewidth]{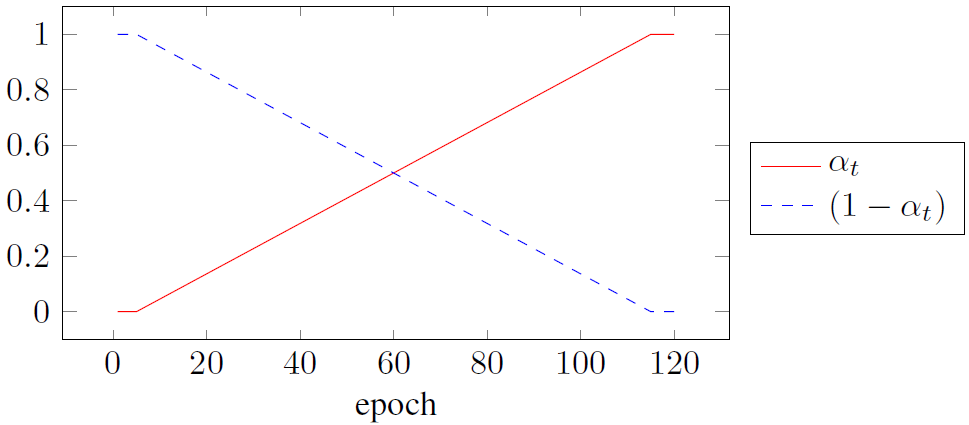}
	\caption{A linear scheduler provides the coefficients of the weighted ensemble. The solid red line denotes the $\alpha_t$ coefficient of the ReLU activation. In the \emph{initial phase} (first 5 epochs) it is set to  0. During the \emph{transition phase} (epochs 6--115) it is progressively increased towards 1. In the \emph{final phase} (last 5 epochs) it remains 1. The dashed blue line denotes the coefficient of the SOTA activation of the ensemble, which is $1-\alpha_t$, and in the \emph{transition phase} it is decayed from 1 towards 0. It remains 0 in the \emph{final phase}, which means that only the ReLUs are used in the final network.}
	\label{fig:schedule}
\end{figure}

In the first ensemble variant we create a simple weighted ensemble model linearly combining the two activations, \emph{i.e.}
\begin{equation}\label{eq:lin_ens}
f_{w}\left(x,\alpha\right) := \alpha \cdot \relu\left(x\right) + \left(1-\alpha\right) \cdot \acts\left(x\right),
\end{equation}
where $\acts$ denotes one of the SOTA activations, \eg in our experiments we use $\acts\in \left\{\gelu,\swish,\mish\right\}$. $f_{w}$ is parameterized by $\alpha$, which controls the contribution of the two activations. Note that parameter $\alpha$ is not learned, but is progressively updated during training as follows. In the \emph{initial phase} we set $\alpha=0$, which means that the ReLU activation is disabled. Next, in the \emph{transition phase} we slowly update $\alpha$ towards $1$, such that the network is progressively adapted to the ReLU activation. In our experiments we use a linear schedule to update $\alpha$, \emph{i.e.}
\begin{equation}\label{eq:schedule}
\alpha_{t} = \frac{t}{T_{trans}} ,
\end{equation} 
where $t$ denotes the t\textit{th} iteration of the \emph{transition phase}, and $T_{trans}$ is the total number of iterations of this phase. As the training progresses ($t \to T_{trans}$) the $\alpha_t$ coefficient of the ReLU in \cref{eq:lin_ens} gets higher ($\alpha \to 1$). Finally, in the \emph{final phase} of PEA, the training is finished with $\alpha=1$, \emph{i.e.} only the ReLU activation is used. This linear schedule is demonstrated in \cref{fig:schedule}, where the coefficient of the ReLU activation (\emph{i.e. } $\alpha_t$) is denoted by a solid red line, and the coefficient of the SOTA activation is by a dashed blue line.

\subsection{Stochastic Ensemble}\label{sec:stoch_ens}

In the second variant we create an ensemble, where the activation is randomly selected. In our experiments the ensemble is composed of two activations, and they are sampled according to a Bernoulli distribution, where the parameter of the distribution is updated such that as the training progresses, the ReLU activation is selected with increasing probability. Formally, we define the stochastic ensemble as
\begin{equation}\label{eq:stoch_ens}
f_{s}\left(x,\alpha\right) := 
\left\{
\begin{array}{ll}
\acts\left(x\right)&\mathrm{if\ } r=0\\
\relu\left(x\right)&\mathrm{if\ } r=1
\end{array}
\right.,r\sim\mathrm{Bern}\left(\alpha\right)
\end{equation}
We apply the same scheduling as used with the weighted ensemble in \cref{sec:lin_ens}, we start with $\alpha=0$ in the \emph{initial phase} ($r=0$, only $\acts$ is sampled), followed by the \emph{transition phase} using the parameter schedule of \cref{eq:schedule}, and finally using $\alpha=1$ in the \emph{final phase}, where $r=1$ means that only the $\relu$ activation is used. Note that this randomized mechanism used in $f_{s}$ is similar to the Dropout \cite{Srivastava2014}, where tensor elements are randomly zeroed, \emph{i.e.} Dropout could be formulated by using $\acts\left(x\right)=0$ in \cref{eq:stoch_ens}.

\section{Experiments}
For the experiments we implemented the proposed PEA methods in TensorFlow: the two ensemble activations discussed in \cref{sec:lin_ens,sec:stoch_ens} as Keras Layers, and the ensemble parameter schedulers of \cref{eq:schedule} as Keras Callbacks.

\subsection{ImageNet Classification}\label{sec:exp_class}
First, in ImageNet classification we evaluated four compact neural networks, including AlexNet \cite{Krizhevsky2012}, ResNet-18 \cite{He2016}, ERFNet \cite{Romera2018}, and MobileNet \cite{Howard2017}. In AlexNet we replaced the local response normalization (LRN) with batch normalization (BN) \cite{Ioffe2015} as proposed in \cite{Simon2016}, referred as AlexNet-BN in the rest of the paper. For the other networks we used their original architecture without any modifications. As the details of the architecture of the ERFNet ImageNet classifier was not discussed in the published paper \cite{Romera2018}, similarly to ResNet-18 we applied global average pooling on the last feature map to extract a feature vector, and used a single fully-connected layer for classification. Moreover, we use the smallest MobileNet variant, which has a width multiplier of $\alpha=0.25$, referred as MobileNet-0.25 in the rest of the paper.

In our experiments we used the Image Classification repository from the TensorFlow Model Garden\footnote{\label{fn:tfgarden}\url{https://github.com/tensorflow/models/tree/master/official/vision/image_classification}}. Our training setup is based on the default settings of the ResNet-50 configuration: crop and flip augmentations, ImageNet standardized input, batch size of 256, label smoothing with $\alpha=0.1$, SGD optimizer with momentum 0.9, a piece-wise learning rate schedule reducing the learning rate by a factor of 0.1 three times (at epochs 30, 60, and 80), and using a linear warm-up period of 5 epochs. However, we increased the training epochs from 90 to 120, which slightly increased the performance of the ResNet baseline. We used these settings in all our experiments, and additional network specific parameters are listed below.

\textbf{AlexNet-BN}: The initial learning rate is set to 0.05, the weight decay is 1e-4, and we apply Dropout with rate 0.5 before the first and second fully-connected layers.

\textbf{ResNet-18}: The initial learning rate is set to 0.1, the weight decay is 1e-4, and no Dropout layer is used.

\textbf{ERFNet}: The initial learning rate is set to 0.1, the weight decay is 1e-4, and we apply Dropout with rate 0.1 in all 1D-factorized non-bottleneck residual modules.

\textbf{MobileNet-0.25}: The initial learning rate is set to 0.1, the weight decay is 4e-5 (no decay for the depthwise convolutions), and we apply Dropout with rate 1e-3 before the fully-connected layer.

First, we trained the original models to reproduce the baselines. MobileNet-0.25 uses the ReLU6 activation by default. Therefore, first we re-trained this model with ReLU6 and obtained similar results as reported in the original publication, \emph{i.e.} our MobileNet-0.25 with ReLU6 had 50.4\% top-1 accuracy, while 50.6\% was reported in \cite{Howard2017}. Next, we replaced ReLU6 with ReLU, and in our experiments this ReLU variant is considered as baseline. \Cref{table:baselines} summarizes our ReLU baseline networks, compared to the results available from the literature. The ResNet-18 accuracy is from the public repository provided by the authors\footnote{\label{fn:resnet}\url{https://github.com/facebookarchive/fb.resnet.torch/blob/master/pretrained/README.md}}.
\begin{table}
	\caption{Top-1 accuracy of the reproduced baseline ImageNet classifiers using ReLU. Our reported results are the average of 5 runs.}
	\label{table:baselines}
	\begin{center}
		\begin{tabular}{lrr}
			\toprule
			\textbf{Baseline ReLU net} & \textbf{Top-1 (repr)} & \textbf{Top-1 (publ)} \\
			\midrule
			AlexNet-BN \cite{Simon2016} & 60.74 & 60.1 \\
			ResNet-18 \cite{He2016} & 70.31 & 69.57 \\
			ERFNet \cite{Romera2018} & 66.35 & N/A \\
			MobileNet-0.25 \cite{Howard2017} & 51.17 & N/A\\
			\hline
		\end{tabular}
	\end{center}
\end{table}

Next, we replaced the ReLU activation with SOTA activation functions and repeated the evaluation process, using the same settings. We considered this modified network as our \emph{Upper limit}, which improves the performance by using a more efficient activation. Note that although in some cases it is recommended to tune the training settings (\eg compared to ReLU networks a lower learning rate was recommended for Swish \cite{Ramachandran2018}), we did not modify any hyper-parameters because our goal was not to tune the network to the highest top-1 score, but to see the improvements on the ReLU network induced by applying the proposed method.

Finally, we trained the ReLU networks using the proposed PEA training. In all cases the \emph{transition phase} started at the 5th epoch, and for the duration of the \emph{final phase} we evaluated two settings. In the first setup the \emph{final phase} included the last 5 epochs (epochs 116--120) only, while in the second setup we increased its duration and used the last 30 epochs (epochs 91--120).

Our ImageNet classification results using the stochastic and weighted variants of PEA are presented in \cref{table:stochresult,table:weightresult} respectively. In general, the stochastic ensemble variant improved the accuracy of all networks in all configurations (see \cref{table:stochresult}). On the other hand, the weighted ensemble (see \cref{table:weightresult}) showed superior performance with AlexNet-BN, but there were some failure cases with the other networks. We discuss our network specific findings below.

\begin{table*}
	\caption{ImageNet top-1 accuracy of the ImageNet classifiers using the \emph{stochastic} variant of PEA, the reported results are the average of 5 runs, and all networks are trained for 120 epochs. We report the results of trainings with both the longer and the shorter \emph{transition phase}, ending at epochs 115 and 90 respectively. All trained networks use ReLU activations only.}
	\label{table:stochresult}
	\begin{center}
		\begin{tabular}{lrrrrrrr}
			\toprule
			\textbf{Network} & \textbf{Baseline} & \multicolumn{2}{c}{\textbf{PEA-GELU}} & \multicolumn{2}{c}{\textbf{PEA-Swish}} & \multicolumn{2}{c}{\textbf{PEA-Mish}} \\
			\cmidrule(l){3-8} 
			& & \multicolumn{1}{c}{115} & \multicolumn{1}{c}{90} & \multicolumn{1}{c}{115} & \multicolumn{1}{c}{90} & \multicolumn{1}{c}{115} & \multicolumn{1}{c}{90} \\
			\midrule 
			AlexNet-BN     & 60.74\% & 61.19\% & \underline{61.29\%} & 61.21\% & 61.02\% & 61.14\% & 61.15\% \\
			ResNet-18      & 70.31\% & \underline{70.51\%} & 70.48\% & 70.40\% & 70.49\% & 70.41\% & 70.38\% \\
			ERFNet         & 66.35\% & 66.61\% & \underline{66.70\%} & 66.65\% & 66.70\% & 66.62\% & 66.60\% \\
			MobileNet-0.25 & 51.17\% & 51.62\% & 51.71\% & 51.46\% & 51.55\% & 51.73\% & \underline{51.77\%} \\
			\bottomrule 
		\end{tabular}
	\end{center}
\end{table*}

\begin{table*}
	\caption{ImageNet top-1 accuracy of the ImageNet classifiers using the \emph{weighted} variant of PEA, the reported results are the average of 5 runs, and all networks are trained for 120 epochs. We report the results of trainings with both the longer and the shorter \emph{transition phase}, ending at epochs 115 and 90 respectively. All trained networks use ReLU activations only.}
	\label{table:weightresult}
	\begin{center}
		\begin{tabular}{lrrrrrrr}
			\toprule
			\textbf{Network} & \textbf{Baseline} & \multicolumn{2}{c}{\textbf{PEA-GELU}} & \multicolumn{2}{c}{\textbf{PEA-Swish}} & \multicolumn{2}{c}{\textbf{PEA-Mish}} \\
			\cmidrule(l){3-8} 
			& & \multicolumn{1}{c}{115} & \multicolumn{1}{c}{90} & \multicolumn{1}{c}{115} & \multicolumn{1}{c}{90} & \multicolumn{1}{c}{115} & \multicolumn{1}{c}{90} \\
			\midrule 
			AlexNet-BN     & 60.74\% & 61.54\% & \underline{61.57\%} & 61.49\% & 61.30\% & 61.35\% & 61.48\% \\
			ResNet-18      & 70.31\% & 70.45\% & \underline{70.51\%} & 70.25\% & 70.38\% & 70.17\% & 70.35\% \\
			ERFNet         & 66.35\% & 66.54\% & 66.59\% & 66.55\% & \underline{66.69\%} & 66.37\% & 66.59\% \\
			\bottomrule 
		\end{tabular}
	\end{center}
\end{table*}

\textbf{AlexNet-BN}. All of the evaluated PEA settings improved the accuracy of AlexNet-BN. In general, the weighted ensemble was clearly superior, and overall the GELU activation performed slightly better.

\textbf{ResNet-18}. Both ensemble methods performed equally well with the GELU activation. In general, the stochastic ensemble variant was clearly superior for both the Swish and the Mish activations, all these configurations improved the performance. On the other hand, two weighted ensembles resulted in performance degradation, a 0.14\% and a 0.06\% drop in top-1 accuracy, when the longer \emph{transition phase} was used. In general, the shorter \emph{transition phase} was beneficial for the weighted variant.

\textbf{ERFNet}. All PEA configurations improved the accuracy. In general, the stochastic ensemble variant was slightly better, and its performance gain was very similar in all settings. Overall the shorter \emph{transition phase} was superior in both ensemble variants. 

\textbf{MobileNet-0.25}. The stochastic ensemble improved the performance in all settings, where Mish worked the best, GELU performed slightly worse, and Swish was clearly inferior. However, with the weighted ensemble variants we observed sudden degradation near the end of the \emph{transition phase}. We are planning to investigate this issue in the future, but as this phenomenon was not present with other networks, we suspect that it might be related to the compact depthwise convolutions used in MobileNet.
We also tried the weighted ensemble variant with the ELU activation \cite{Clevert2016}, and surprisingly we did not observe this phenomenon, and the accuracy increased to 51.29\%. Nevertheless, this improvement is inferior as compared to the stochastic variant.

In \cref{table:result} our results are summarized and are compared to the ReLU baseline network and to the network using SOTA activation (denoted by \emph{Upper limit}). The PEA configurations in the bracket denote the training setup starting with the ensemble either \textbf{W}eighted or \textbf{S}tochastic, followed by the activation function \textbf{G}ELU, \textbf{S}wish, or \textbf{M}ish. In our evaluations we selected the best performing model from the \emph{final phase} of PEA, \emph{i.e.} where the network uses the ReLU activation only.

\begin{table}
	\caption{Top-1 accuracy of the ImageNet classifiers, the reported results are the average of 5 runs. The training setup of our proposed PEA trained ReLU networks are indicated in the brackets, \emph{i.e.} the first character indicates the ensemble method (\textbf{W}eighted or \textbf{S}tochastic), the second is the activation used in the ensemble (\textbf{G}ELU, \textbf{S}wish, or \textbf{M}ish). The end epoch of the \emph{transition phase} is 90 in all cases. The \emph{Upper limit} denotes the best performing network using SOTA activation.}
	\label{table:result}
	\begin{center}
		\begin{tabular}{lrrr}
			\toprule
			\textbf{Network} & \textbf{ReLU} & \textbf{ReLU PEA} & \textbf{Upper limit} \\
			\midrule 
			AlexNet-BN & 60.74 & 61.57 (WG) & 62.35 (G) \\
			ResNet-18 & 70.31 & 70.51 (WG) & 70.89 (S) \\
			ERFNet & 66.35 & 66.70 (SG) & 67.10 (M) \\
			MobileNet-0.25 & 51.17 & 51.77 (SM) & 53.40 (S)\\
			\bottomrule 
		\end{tabular}
	\end{center}
\end{table}

\subsubsection{Improving Large Networks}
Although our main goal was to improve the performance of compact ReLU networks, we also evaluated the proposed methods on a larger ResNet-50 network. Apart from increasing the training epochs from 90 to 120, we did not modify the ResNet-50 configuration in the Image Classification repository of the TensorFlow Model Garden\cref{fn:tfgarden}. First, we trained our ReLU baseline five times, and our average top-1 accuracy was 76.81\%, while 75.99\% is reported in the public repository of the authors\cref{fn:resnet}. Next, we trained the network with PEA, and we successfully improved the accuracy to 77.04\% using the stochastic ensemble and the Swish activation. We observed performance gain with both the weighted and the stochastic variants, and the latter performed slightly better. In general, the Swish activation was clearly superior in both ensemble variants. Mish performed the worst in all configurations, and in one case it decreased the accuracy by 0.05\%. Overall the shorter \emph{transition phase} worked slightly better for both ensemble variants. We summarize our results in \cref{table:rn50result}.
\begin{table*}[t]
	\caption{ImageNet top-1 accuracy improvements of the ResNet-50 networks using the PEA training, the reported results are the average of 5 runs, and all networks are trained for 120 epochs. We report the results of trainings with both the longer and the shorter \emph{transition phase}, ending at epochs 115 and 90 respectively. All trained networks use ReLU activations only.}
	\label{table:rn50result}
	\begin{center}
		\begin{tabular}{lrrrrrr}
			\toprule
			\textbf{Method} & \multicolumn{2}{c}{\textbf{PEA-GELU}} & \multicolumn{2}{c}{\textbf{PEA-Swish}} & \multicolumn{2}{c}{\textbf{PEA-Mish}} \\
			\cmidrule(l){2-7} 
			& \multicolumn{1}{c}{115} & \multicolumn{1}{c}{90} & \multicolumn{1}{c}{115} & \multicolumn{1}{c}{90} & \multicolumn{1}{c}{115} & \multicolumn{1}{c}{90} \\
			\midrule 
			Weighted     & +0.08\% & +0.18\% & +0.13\% & +0.21\% & -0.05\% & +0.12\% \\
			Stochastic   & +0.15\% & +0.21\% & \underline{+0.23\%} & +0.22\% & +0.13\% & +0.20\% \\
			\bottomrule 
		\end{tabular}
	\end{center}
\end{table*}

\subsection{Cityscapes Segmentation}\label{sec:exp_semseg}

We demonstrate our proposed methods on semantic segmentation, where the task is to classify each pixel into a set of categories. In our experiments we used the compact ERFNet segmentation network \cite{Romera2018}, and evaluated on the Cityscapes dataset \cite{Cordts2016} using the suggested 19 semantic categories. The authors proposed several tricks to boost the performance of their final model, \eg pre-training the backbone on ImageNet, a training procedure composed of multiple stages, and tuning the class weights in the loss function. We decided not to use any of these tricks, because we were interested in the performance gain obtained by using the proposed PEA training. Therefore, we performed single-stage from-scratch trainings using random initialized weights, standard cross-entropy loss, and an input size of $512\times 256$. To train our networks we slightly modified the settings of \cite{Romera2018}: batch size is increased to 25, and the initial learning rate to 1e-3, we use weight decay 1e-4, but we kept the dropout rate 0.3 proposed by the authors. We trained the models for 500 epochs from scratch, using the RAdam optimizer \cite{Liu2020} and a cosine learning rate schedule with a linear warm-up period of one epoch. For data augmentation we used random translation and horizontal flip as proposed by the authors. Similarly to our ImageNet experiments, we also repeated the Cityscapes experiments five times, and computed their average mIOU score. With this setup we achieved mIOU=58.26\% on the validation set, which is comparable to the result of mIOU=58.37\% reported in \cite{Romera2018} using the same input resolution (see Table III in their publication).

Next, we replaced the ReLU with the SOTA activations, and repeated our evaluations, while keeping other parameters intact. The GELU and Mish activations increased the average performance to mIOU 58.62\% and 58.84\% respectively, but with the Swish we observed performance degradation of mIOU 57.88\%, so we decided to remove Swish from further experiments. This improved mIOU score is considered as the \emph{Upper limit}, which can be obtained by simply replacing the ReLU with a better activation function.

Finally, we evaluated the proposed PEA training method using GELU and Mish. As for the parameter scheduling we evaluated two settings for the duration of the \emph{final phase}, where the network contains ReLU activations only. In the first setup this phase consists of the last 50 epochs, while in the second one it is increased to 150 epochs. With the proposed PEA training we successfully improved the performance of the network by 0.34\% mIOU. In general, the weighted ensemble variant was superior, and it improved the performance in all settings. In \cref{table:cityresult} we compare our best performing model to the ReLU baseline and to the \emph{Upper limit} model using the Mish activation.
\begin{table}[ht]
	\caption{Performance of the ERFNet segmentation network on the Cityscapes validation set, the reported results are the average of 5 runs. We used single-stage trainings, and random initialized weights. The training setup of the proposed PEA trained ReLU networks are indicated in the brackets, \emph{i.e.} the first character indicates the ensemble method (\textbf{W}eighted or \textbf{S}tochastic), the second is the activation used in the ensemble (\textbf{G}ELU or \textbf{M}ish), and the number denotes the end epoch of the \emph{transition phase}. The \emph{Upper limit} denotes the best performing network using SOTA activation.}
	\label{table:cityresult}
	\begin{center}
		\begin{tabular}{lr}
			\toprule
			\textbf{Network} & \textbf{mIOU} \\
			\midrule 
			ReLU baseline & 58.26\% \\
			ReLU with PEA (WM450) & 58.60\% \\
			Upper limit (M) & 58.84\% \\
			\bottomrule 
		\end{tabular}
	\end{center}
\end{table}

\section{Conclusions}

In this paper we proposed Progressive Ensemble Activation or PEA, a training method to improve the accuracy of ReLU networks \emph{for free}. During network training PEA creates an ensemble activation composed of the ReLU and an efficient SOTA activation (\eg GELU, Swish, Mish). The coefficients of the ensemble are progressively updated by a scheduler, such that initially only the SOTA activation is used, and then the network is slowly adapted to the ReLU activation in a \emph{transition phase}, where its coefficient is increased. Finally, network training is finished with ReLUs only, and the other SOTA activations are disabled. This means that in the final network the ensemble activation can be replaced by a ReLU, and the network can be executed in environments where only the ReLU is supported, such as embedded AI accelerators. First, we evaluated PEA on the ImageNet classification task, and we applied it to four different compact network architectures using three SOTA activation functions. According to our benchmark, the improvements are in the range of 0.2--0.8\% top-1 accuracy. We also applied our method to a larger ResNet-50 network, and improved its top-1 performance by 0.23\%. Finally, we applied the proposed method to improve the performance of a compact segmentation network, and on Cityscapes we observed a performance improvement of 0.34\% mIOU. The advantage of the proposed method is that both ensemble variants are relatively simple, they do not increase the training time and the memory footprint drastically, which means the method can also be applied to improve larger networks. In our experiments we used training parameters (\eg learning rate) which were tuned to the original ReLU network. As these are usually not optimal for other SOTA activations, the proposed method might be further extended such that it takes these differences into account, \eg by progressively adapting the learning rate, or using a Swish parameter optimized for a given network architecture, or by using alternative scheduling of the coefficients. Another possible research direction would be to use the two ensemble variants jointly, for example by selecting the optimal setup for each network layer separately. As the proposed method is complementary to other previous \emph{for-free} methods, it would also be interesting to see the benefits of combining them during training. 

{\small
\bibliographystyle{ieee_fullname}
\bibliography{PaperForReview}

\begin{thebibliography}{10}\itemsep=-1pt

\bibitem{Cho2019}
Jang~Hyun Cho and Bharath Hariharan.
\newblock On the efficacy of knowledge distillation.
\newblock In {\em Proceedings of the {IEEE} International Conference on
  Computer Vision}, pages 4793--4801, 2019.

\bibitem{Clevert2016}
Djork{-}Arn{\'{e}} Clevert, Thomas Unterthiner, and Sepp Hochreiter.
\newblock Fast and accurate deep network learning by exponential linear units
  ({ELUs}).
\newblock In {\em Proceedings of the 4th International Conference on Learning
  Representations}, 2016.

\bibitem{Cordts2016}
Marius Cordts, Mohamed Omran, Sebastian Ramos, Timo Rehfeld, Markus Enzweiler,
  Rodrigo Benenson, Uwe Franke, Stefan Roth, and Bernt Schiele.
\newblock The {Cityscapes} dataset for semantic urban scene understanding.
\newblock In {\em Proceedings of the IEEE Conference on Computer Vision and
  Pattern Recognition}, 2016.

\bibitem{Cubuk2020}
Ekin~D. Cubuk, Barret Zoph, Jonathon Shlens, and Quoc~V. Le.
\newblock {RandAugment}: Practical automated data augmentation with a reduced
  search space.
\newblock In {\em Proceedings of the 34th Conference on Neural Information
  Processing Systems}, June 2020.

\bibitem{Deng2009}
Jia Deng, Wei Dong, Richard Socher, Li{-}Jia Li, Kai Li, and Li Fei{-}Fei.
\newblock {ImageNet}: A large-scale hierarchical image database.
\newblock In {\em Proceedings of the IEEE Conference on Computer Vision and
  Pattern Recognition}, pages 248--255, 2009.

\bibitem{Dosovitskiy2021}
Alexey Dosovitskiy, Lucas Beyer, Alexander Kolesnikov, Dirk Weissenborn,
  Xiaohua Zhai, Thomas Unterthiner, Mostafa Dehghani, Matthias Minderer, Georg
  Heigold, Sylvain Gelly, Jakob Uszkoreit, and Neil Houlsby.
\newblock An image is worth 16x16 words: Transformers for image recognition at
  scale.
\newblock In {\em Proceedings of the 9th International Conference on Learning
  Representations}, 2021.

\bibitem{Elfwing2018}
Stefan Elfwing, Eiji Uchibe, and Kenji Doya.
\newblock Sigmoid-weighted linear units for neural network function
  approximation in reinforcement learning.
\newblock {\em Neural Networks}, 107:3--11, November 2018.

\bibitem{Ghiasi2021}
Golnaz Ghiasi, Yin Cui, Aravind Srinivas, Rui Qian, Tsung{-}Yi Lin, Ekin~D.
  Cubuk, Quoc~V. Le, and Barret Zoph.
\newblock Simple copy-paste is a strong data augmentation method for instance
  segmentation.
\newblock In {\em Proceedings of the {IEEE} Conference on Computer Vision and
  Pattern Recognition}, pages 2918--2928, 2021.

\bibitem{Glorot2011}
Xavier Glorot, Antoine Bordes, and Yoshua Bengio.
\newblock Deep sparse rectifier neural networks.
\newblock In {\em Proceedings of the 14th International Conference on
  Artificial Intelligence and Statistics}, pages 315--323, 2011.

\bibitem{Guo2020}
Shuxuan Guo, Jose~M. Alvarez, and Mathieu Salzmann.
\newblock {ExpandNets}: Linear over-parameterization to train compact
  convolutional networks.
\newblock In {\em Proceedings of the 34th Conference on Neural Information
  Processing Systems}, pages 1298--1310, 2020.

\bibitem{He2015}
Kaiming He, Xiangyu Zhang, Shaoqing Ren, and Jian Sun.
\newblock Delving deep into rectifiers: Surpassing human-level performance on
  {ImageNet} classification.
\newblock In {\em Proceedings of the IEEE International Conference on Computer
  Vision}, pages 1026--1034, 2015.

\bibitem{He2016}
Kaiming He, Xiangyu Zhang, Shaoqing Ren, and Jian Sun.
\newblock Deep residual learning for image recognition.
\newblock In {\em Proceedings of the IEEE Conference on Computer Vision and
  Pattern Recognition}, pages 770--778, 2016.

\bibitem{Hendrycks2016}
Dan Hendrycks and Kevin Gimpel.
\newblock Gaussian error linear units ({GELUs}).
\newblock {\em arXiv preprint arXiv:1606.08415}, 2016.

\bibitem{Hinton2015}
Geoffrey Hinton, Oriol Vinyals, and Jeffrey Dean.
\newblock Distilling the knowledge in a neural network.
\newblock In {\em NIPS Deep Learning and Representation Learning Workshop},
  2015.

\bibitem{Howard2017}
Andrew~G. Howard, Menglong Zhu, Bo Chen, Dmitry Kalenichenko, Weijun Wang,
  Tobias Weyand, Marco Andreetto, and Hartwig Adam.
\newblock {MobileNets}: Efficient convolutional neural networks for mobile
  vision applications.
\newblock {\em CoRR}, abs/1704.04861, 2017.

\bibitem{Huang2016}
Gao Huang, Yu Sun, Zhuang Liu, Daniel Sedra, and Kilian~Q. Weinberger.
\newblock Deep networks with stochastic depth.
\newblock In {\em Proceedings of the 14th European Conference on Computer
  Vision}, pages 646--661, 2016.

\bibitem{Ioffe2015}
Sergey Ioffe and Christian Szegedy.
\newblock Batch normalization: Accelerating deep network training by reducing
  internal covariate shift.
\newblock In {\em Proceedings of the 32nd International Conference on Machine
  Learning}, pages 448--456, 2015.

\bibitem{Podoprikhin2018}
Pavel Izmailov, Dmitrii Podoprikhin, Timur Garipov, Dmitry~P. Vetrov, and
  Andrew~Gordon Wilson.
\newblock Averaging weights leads to wider optima and better generalization.
\newblock In {\em Proceedings of the 34th Conference on Uncertainty in
  Artificial Intelligence}, 2018.

\bibitem{Kingma2015}
Diederik~P. Kingma and Jimmy Ba.
\newblock Adam: {A} method for stochastic optimization.
\newblock In {\em Proceedings of the 3rd International Conference on Learning
  Representations}, 2015.

\bibitem{Klambauer2017}
Günter Klambauer, Thomas Unterthiner, Andreas Mayr, and Sepp Hochreiter.
\newblock Self-normalizing neural networks.
\newblock In {\em Proceedings of the 31st International Conference on Neural
  Information Processing Systems}, pages 971--980, 2017.

\bibitem{Komodakis2017}
Nikos Komodakis and Sergey Zagoruyko.
\newblock Paying more attention to attention: {Improving} the performance of
  convolutional neural networks via attention transfer.
\newblock In {\em Proceedings of the 5th International Conference on Learning
  Representations}, 2017.

\bibitem{Krizhevsky2012}
Alex Krizhevsky, Ilya Sutskever, and Geoffrey~E. Hinton.
\newblock {ImageNet} classification with deep convolutional neural networks.
\newblock In {\em Proceedings of the 25th International Conference on Neural
  Information Processing Systems}, pages 1097--1105, 2012.

\bibitem{Lin2014}
Tsung{-}Yi Lin, Michael Maire, Serge Belongie, Lubomir Bourdev, Ross Girshick,
  James Hays, Pietro Perona, Deva Ramanan, C.~Lawrence Zitnick, and Piotr
  Doll\'ar.
\newblock {Microsoft COCO}: Common objects in context.
\newblock In {\em Proceedings of the 13th European Conference on Computer
  Vision}, pages 740--755, 2014.

\bibitem{Liu2020}
Liyuan Liu, Haoming Jiang, Pengcheng He, Weizhu Chen, Xiaodong Liu, Jianfeng
  Gao, and Jiawei Han.
\newblock On the variance of the adaptive learning rate and beyond.
\newblock In {\em Proceedings of the 8th International Conference on Learning
  Representations}, 2020.

\bibitem{Liu2022}
Ze Liu, Han Hu, Yutong Lin, Zhuliang Yao, Zhenda Xie, Yixuan Wei, Jia Ning, Yue
  Cao, Zheng Zhang, Li Dong, Furu Wei, and Baining Guo.
\newblock Swin transformer {V2}: {Scaling} up capacity and resolution.
\newblock In {\em Proceedings of the IEEE Conference on Computer Vision and
  Pattern Recognition}, 2022.

\bibitem{Liu2021}
Ze Liu, Yutong Lin, Yue Cao, Han Hu, Yixuan Wei, Zheng Zhang, Stephen Lin, and
  Baining Guo.
\newblock Swin transformer: Hierarchical vision transformer using shifted
  windows.
\newblock {\em CoRR}, abs/2103.14030, 2021.

\bibitem{Loshchilov2017}
Ilya Loshchilov and Frank Hutter.
\newblock {SGDR}: Stochastic gradient descent with warm restarts.
\newblock In {\em Proceedings of the 5th International Conference on Learning
  Representations}, 2017.

\bibitem{Lu2020}
Lu Lu, Yeonjong Shin, Yanhui Su, and George Karniadakis.
\newblock Dying {ReLU} and initialization: Theory and numerical examples.
\newblock {\em Communications in Computational Physics}, 28:1671--1706, 11
  2020.

\bibitem{Maas2013}
Andrew~L. Maas, Awni~Y. Hannun, and Andrew~Y. Ng.
\newblock Rectifier nonlinearities improve neural network acoustic models.
\newblock In {\em Proceedings of the International Conference on Machine
  Learning}, 2013.

\bibitem{Mirzadeh2020}
Seyed{-}Iman Mirzadeh, Mehrdad Farajtabar, Ang Li, Nir Levine, Akihiro
  Matsukawa, and Hassan Ghasemzadeh.
\newblock Improved knowledge distillation via teacher assistant: Bridging the
  gap between student and teacher.
\newblock {\em Proceedings of the 34th {AAAI} Conference on Artificial
  Intelligence}, 2020.

\bibitem{Misra2020}
Diganta Misra.
\newblock Mish: {A} self regularized non-monotonic neural activation function.
\newblock In {\em Proceedings of the 31st British Machine Vision Conference},
  2020.

\bibitem{Nair2010}
Vinod Nair and Geoffrey~E. Hinton.
\newblock Rectified linear units improve restricted {Boltzmann} machines.
\newblock In {\em Proceedings of the 27th International Conference on Machine
  Learning}, pages 807–--814, 2010.

\bibitem{Pham2021}
Hieu Pham, Zihang Dai, Qizhe Xie, and Quoc~V. Le.
\newblock Meta pseudo labels.
\newblock In {\em Proceedings of the IEEE Conference on Computer Vision and
  Pattern Recognition}, pages 11557--11568, June 2021.

\bibitem{Ramachandran2018}
Prajit Ramachandran, Barret Zoph, and Quoc~V. Le.
\newblock Searching for activation functions.
\newblock In {\em 6th International Conference on Learning Representations,
  Workshop Track Proceedings}, 2018.

\bibitem{Romera2018}
Eduardo Romera, Jos\'e~M. \'Alvarez, Luis~M. Bergasa, and Roberto Arroyo.
\newblock {ERFNet}: Efficient residual factorized convnet for real-time
  semantic segmentation.
\newblock {\em IEEE Transactions on Intelligent Transportation Systems},
  19(1):263--272, 2018.

\bibitem{Romero2015}
Adriana Romero, Nicolas Ballas, Samira~Ebrahimi Kahou, Antoine Chassang, Carlo
  Gatta, and Yoshua Bengio.
\newblock {FitNets}: Hints for thin deep nets.
\newblock In {\em Proceedings of the 3rd International Conference on Learning
  Representations}, 2015.

\bibitem{Shen2021}
Zhiqiang Shen, Zechun Liu, Dejia Xu, Zitian Chen, Kwang{-}Ting Cheng, and
  Marios Savvides.
\newblock Is label smoothing truly incompatible with knowledge distillation: An
  empirical study.
\newblock In {\em Proceedings of the 9th International Conference on Learning
  Representations}, 2021.

\bibitem{Shen2020}
Zhiqiang Shen and Marios Savvides.
\newblock {MEAL} {V2}: Boosting vanilla {ResNet-50} to 80{\%}+ top-1 accuracy
  on {ImageNet} without tricks.
\newblock {\em CoRR}, abs/2009.08453, 2020.

\bibitem{Simon2016}
Marcel Simon, Erik Rodner, and Joachim Denzler.
\newblock {ImageNet} pre-trained models with batch normalization.
\newblock {\em CoRR}, abs/1612.01452, 2016.

\bibitem{Srivastava2014}
Nitish Srivastava, Geoffrey~E. Hinton, Alex Krizhevsky, Ilya Sutskever, and
  Ruslan Salakhutdinov.
\newblock Dropout: A simple way to prevent neural networks from overfitting.
\newblock {\em Journal of Machine Learning Research}, 15(56):1929--1958, 2014.

\bibitem{Yuan2021}
Lu Yuan, Dongdong Chen, Yi-Ling Chen, Noel Codella, Xiyang Dai, Jianfeng Gao,
  Houdong Hu, Xuedong Huang, Boxin Li, Chunyuan Li, Ce Liu, Mengchen Liu,
  Zicheng Liu, Yumao Lu, Yu Shi, Lijuan Wang, Jianfeng Wang, Bin Xiao, Zhen
  Xiao, Jianwei Yang, Michael Zeng, Luowei Zhou, and Pengchuan Zhang.
\newblock Florence: {A} new foundation model for computer vision.
\newblock {\em arXiv preprint arXiv:2111.11432}, 2021.

\bibitem{Zhang2022}
Hao Zhang, Feng Li, Shilong Liu, Lei Zhang, Hang Su, Jun Zhu, Lionel~M. Ni, and
  Heung{-}Yeung Shum.
\newblock {DINO}: {DETR} with improved denoising anchor boxes for end-to-end
  object detection.
\newblock {\em arXiv preprint arXiv:2203.03605}, 2022.

\bibitem{Zhang2019}
Michael~R. Zhang, James Lucas, Jimmy Ba, and Geoffrey~E. Hinton.
\newblock Lookahead optimizer: k steps forward, 1 step back.
\newblock In {\em Proceedings of the 33rd Conference on Neural Information
  Processing Systems}, pages 9593--9604, 2019.

\end{thebibliography}
}

\end{document}